\pdfoutput=1

\documentclass[11pt]{article}

\usepackage{acl}

\usepackage{times}
\usepackage{latexsym}

\usepackage[T1]{fontenc}

\usepackage[utf8]{inputenc}

\usepackage{microtype}

\usepackage[autolanguage]{numprint}
\usepackage{csquotes}
\usepackage{graphicx}
\graphicspath{ {./images} }

\usepackage{amsmath} 
\usepackage{multirow} 
\usepackage{subcaption} 
\usepackage{booktabs} 

\usepackage{tikz} 
\usetikzlibrary{tikzmark}
\usetikzlibrary{shapes.geometric, arrows}
\tikzstyle{startstop} = [rectangle, rounded corners, minimum width=3cm, minimum height=1cm,text centered, draw=black, fill=blue!30]
\tikzstyle{arrow} = [thick,->,>=stealth]

\tikzset{every tikzmarknode/.style={align=center, inner sep = 1pt,execute at end node={\vphantom{bg}}}}

\title{Does mBERT Understand Romansh? Evaluating Word Embeddings Using Word Alignment.}

\author{Eyal Liron Dolev \\
Zurich Center of Linguistics -- LiZZ  \\
University of Zurich \\
  \texttt{eyalliron.dolev@uzh.ch} 
}
\begin{document}
\maketitle
\begin{abstract}
  We test similarity-based word alignment models (SimAlign and awesome-align) in combination with word embeddings from mBERT and XLM-R on parallel sentences in German and Romansh. Since Romansh is an unseen language, we are dealing with a zero-shot setting. Using embeddings from mBERT, both models reach an alignment error rate of 0.22, which outperforms fast\_align, a statistical model, and is on par with similarity-based word alignment for seen languages. 
  We interpret these results as evidence that mBERT contains information that can be meaningful and applicable to Romansh.

  To evaluate performance, we also present a new trilingual corpus, which we call the DERMIT (DE-RM-IT) corpus, containing press releases made by the Canton of Grisons in German, Romansh and Italian in the past 25 years. The corpus contains \numprint{4547} parallel documents and approximately \numprint{100000} sentence pairs in each language combination. We additionally present a gold standard for German-Romansh word alignment. 
  The data is available at \url{https://github.com/eyldlv/DERMIT-Corpus}.

\end{abstract}

\section{Introduction}\label{sec:introduction}

Romansh is a Romance language spoken by an ever-diminishing number of speakers (approximately \numprint{40000} as of 2020\footnote{\url{https://www.bfs.admin.ch/bfs/de/home/statistiken/bevoelkerung/sprachen-religionen/sprachen.assetdetail.21344032.html}}) mostly in the Canton of Grisons (henceforth Graubünden). The name \emph{Romansh} is an umbrella term for five dialects (Surselvan, Sutselvan, Surmiran, Puter, Vallader), each having normative grammars and orthography \citep[p.~1]{haiman1992}. In 1982, Rumantsch Grischun, a standard written variety of Romansh was launched, its goal being the harmonization of the five dialects and the creation of a language variant that would be understood by all speakers of Romansh and could be used i.a. by officials, government agencies and in formal writings \citep[p.~5]{haiman1992}.\footnotemark{}

\footnotetext{In our paper, whenever we use the name \emph{Romansh} we refer to the standard variety \emph{Rumantsch Grischun}}

Currently, there are no state-of-the-art NLP tools, such as part-of-speech (PoS) tagging or named entity recognition (NER), for Romansh. 
However, multilingual language models (MLMs) were shown to perform well in various tasks in zero-shot settings and on unseen languages, i.e., for languages not included in the training data. 
\citet{pires-etal-2019-multilingual} showed that fine-tuning mBERT for PoS-tagging on one language generalizes well (over 80\% accuracy) for other languages. 
Similarly, the LASER model, which was pre-trained on 93 languages, obtained strong results for sentence embeddings in 112 languages \citep{artexte-schwenk-2019-laser}.
Since Romansh data, raw as well as annotated, is scarce, leveraging MLMs for annotation tasks and other tasks requiring word embeddings, could provide significant advantages when developing language technology and NLP applications for Romansh, instead of training new models from scratch.

We hypothesize that due to the proven success of MLMs in zero-shot settings and on unseen languages, and due to the similarity of Romansh to other Romance languages that were included in pretraining, MLMs should contain information that is meaningful and applicable to processing Romansh. To support this hypothesis, two conditions should be fulfilled:

Word alignments computed with similarity-based systems (such as SimAlign or awesome-align) using word embeddings taken from MLMs should:
\begin{enumerate}
        \item be on par with statistical models;
        \item be on par with alignments computed using multilingual embeddings for seen language pairs.
\end{enumerate}

If both conditions are fulfilled, we would view this as an indication that these models contain information that can be meaningfully applied to processing Romansh. 

To test our hypothesis, we collected a new trilingual corpus, which is a continuation of the ALLEGRA corpus by \citet{scherrer-cartoni-2012-trilingual} (Section~\ref{sec:corpus}), extracted parallel sentences (Section~\ref{sec:sent-align}) and created a word alignment gold standard by annotating 600 sentences (Section~\ref{sec:gold-standard}). The experiments and the results are described in Section~\ref{sec:experiments}.

\section{Previous Work}
Romansh has enjoyed a considerable amount of attention from the NLP community despite its low number of speakers. This is arguably due to several favorable circumstances: It is spoken in a highly modernized land, it enjoys official legal status as a national language and it is promoted and protected by law.\footnotemark{}  

\footnotetext{Cf.~\emph{\href{https://www.gr-lex.gr.ch/app/de/texts_of_law/110.100}{Verfassung des Kantons Graubündens}} and \emph{\href{https://www.gr-lex.gr.ch/app/de/texts_of_law/492.100}{Sprachengesetz des Kantons Graubündens}}.}

\citet{scherrer-cartoni-2012-trilingual} compiled a trilingual corpus using the press releases published by the canton of Graubünden, \citet{baumgartner-etal-2014-morpho} created a system for automatic morphological analysis, \citet{weibel-2014} compiled two bilingual (German-Romansh), sentence- and word-aligned corpora, containing the cantonal press releases and legal texts. 
These corpora are available on the website \enquote{multilingwis}\footnotemark{} \citep{multilingwis}. 
Romansh was also part of the multilingual corpus used by \citet{volk-clematide-2014-detecting} for evaluating code-switching detection and it was used by \citet{muller-etal-2020-domain} for evaluating the performance of out-of-domain machine translation.
\footnotetext{\url{https://pub.cl.uzh.ch/projects/sparcling/multilingwis2.demo/}}
Last year, in 2022, TextShuttle, a Zurich-based company specializing in machine translation, developed and released a machine translation system for Romansh (translating to or from German, French, Italian and English).


\section{Corpus}\label{sec:corpus}
Our corpus is based on public press releases published by the Canton of Graubünden.

The Canton of Graubünden is the only Swiss canton with three official languages: German, spoken mostly in the northern parts of the canton, Italian, spoken mostly in the south, e.g., in the valleys Poschiavo and Mesolcina, and Romansh, spoken sporadically in several valleys, e.g., Engadine or Domleschg.

The canton has been publicly publishing press releases in these three official languages (for Romansh, the standard variety \enquote{Rumantsch Grischun} is used) at least since 1997, and they are publicly available on the canton's website.
The press releases are short texts, usually consisting of about 3-5 paragraphs, 
which are intended to inform the public about events and occurrences in the canton and in the cantonal authorities, about decisions taken by the cantonal government, etc. Some typical topics are politics, economics and health. 

\subsection{Scraping and Document Alignment}
To collect said press releases and create a parallel corpus, we collected all the press releases published from 2010 until January 2023 on the canton's website.
We then aligned the documents using URL matching. 
The press releases published between 1997 and 2009 cannot be aligned automatically using URL matching. 
However, since \citet{scherrer-cartoni-2012-trilingual} already manually aligned these documents for their ALLEGRA corpus, we included them, as is, in our own corpus. We thus see our work as a continuation of the ALLEGRA corpus.

Table~\ref{tab:corpus} in Appendix~\ref{sec:appendix} gives information about the total number of documents in each language and the number of documents that exist in at least two languages. 

Both the aligned and the unaligned documents are presented as JSON files (one file per year) and as an SQLite database, which both allow for simple and fast queries.

\subsection{Sentence Alignment}\label{sec:sent-align}
For sentence tokenization/segmentation, we used NLTK's \citep{bird-2009-nltk} Punkt tokenizer, which is pre-trained for German and Italian. For Romansh, we used the Italian tokenizer, which we extended with a list of Romansh abbreviations to improve sentence segmentation. 

We then sentence-aligned the press releases using hunalign \citep{hunalign}, a hybrid sentence-alignment system combining a dictionary-based and a length-based method. Although not a requirement, we supplied hunalign with a German-Romansh dictionary\footnotemark{}. We have no gold standard for sentence alignment, but out of 611 sentences extracted for the task of annotating the gold standard, only 11 sentences had to be removed due to misalignment, which corresponds to an alignment error rate of 1.8\%.

\footnotetext{The dictionary was downloaded from \url{https://www.pledarigrond.ch/rumantschgrischun}}

After removing duplicates and obvious misalignments using simple heuristics, we were able to extract  \numprint{106091} German-Romansh sentence pairs, \numprint{103441} German-Italian sentence pairs and \numprint{102757} Romansh-Italian sentence pairs, see also Table~\ref{tab:sentence-pairs}.

\begin{table}
        \centering
        \begin{tabular}{lccc}
                \toprule
                Comb. & Sentences & Tokens & Types\\
                \midrule
                DE-RM & \numprint{106091} & 1.9M, 2.4M & 97K, 51K \\
                DE-IT & \numprint{103441} & 1.8M, 2.2M & 96K, 58K \\
                IT-RM & \numprint{102757} & 2.3M, 2.2M & 50K, 58K \\
                \bottomrule                
        \end{tabular}
        \caption{Unique sentence pairs extracted from the corpus for the three language combinations: German-Romansh, German-Italian and Italian-Romansh.}
        \label{tab:sentence-pairs}
\end{table}

\subsection{Word Alignment}
Word alignments are objects mapping each word in a string of the source language $e$ to the words in a string of the target language $f$ \citep[p.~84]{koehn2009}.  They are traditionally computed using the IBM Models 1-5, presented by \citet{brown-etal-1993-mathematics}, and there are several implementations of these models, e.g., Giza++ by \citet{och-ney-2000-improved}, fast\_align by \citet{dyer-etal-2013-simple} or eflomal by \citet{Ostling2016efmaral}.

Word alignments used to be an integral part of statistical machine translation (SMT), and although they are not necessary for neural machine translation (NMT), they can be incorporated into different techniques to improve NMT (e.g., \citet{alkhouli-etal-2018-alignment} which use explicit alignments for their NMT system). They are also still in use in other subfields of NLP, such as improving bilingual lexicon induction \citep{shi-etal-2021-bilingual} or filtering noisy parallel corpora \citep{kurfali-ostling-2019-noisy}. 
See \citet{steingrimsson-etal-2021-combalign} for a short overview.

More recently, word alignment systems that take advantage of contextualized word embeddings that are extracted from MLMs have been proposed.
These systems compute word alignments by extracting from an MLM the embeddings of words in a sentence pair and putting them in a similarity matrix. The system then computes for each embedding vector representing a word in the source sentence the most similar vector or vectors representing the words in the target sentence, which are then aligned.
The major advantage of similarity-based word alignment is that the alignment quality is not affected by the amount of parallel data. Therefore, high-quality word alignments can be computed also in low-resource settings, when parallel data is scarce \citep{jalili-sabet-etal-2020-simalign}.
Two systems that use this concept are SimAlign by \citet{jalili-sabet-etal-2020-simalign} and awesome-align by \citet{dou-neubig-2021-word}.

\section{Gold Standard for Word Alignment}\label{sec:gold-standard}
In order to evaluate different word-alignment systems, a gold standard is needed \citep[p.~115]{koehn2009}. 
To this end, we created a gold standard consisting of 600 German-Romansh sentence pairs annotated for word alignment. 
The annotations were done using AlignMan, an annotation tool created by \citet{steingrimsson-etal-2021-combalign}.

\subsection{Guidelines}
Since descriptions of gold standard annotations for word alignment are rare in the literature, we decided to elaborate here on the annotation process. 

To ensure consistency, a set of annotation guidelines were formulated. 
Following previous work, we aimed to \enquote{[a]lign as small segments as possible and as long segments as necessary.} (\citet{lines2007} citing \citet{Veronis2000}).
More explicitly, we adhered to three general principles:

\paragraph{Principle I.} \textbf{Align only unambiguous words.} 
Since the annotation tool we used (ManAlign by \citet{steingrimsson-etal-2021-combalign}) does not allow the use of confidence labels, i.e., it is only possible to annotate Sure but not Possible alignments, we only aligned words which are unambiguously mutual translations.

\paragraph{Principle II.} \textbf{Prefer 1-to-1 alignments over 1-to-n alignments or n-to-n alignments. }
Since all alignments are Sure alignments, and since we evaluate word alignment, and not phrase alignment systems, 1-to-n alignments should be avoided.
However, if a single word in the source sentence lexically corresponds to several words in the target sentence, they are aligned.
In case words are repeated in one language but not in the other language, they are aligned only once.

\paragraph{Principle III.} \textbf{Prefer lexical alignment over other types of alignments such as part-of-speech alignments or morphosyntactical alignemnts.} 
Only words that are translations of each other also outside the specific context of the sentence pair should be aligned. 
In cases of paraphrasing during translations, words should remain unaligned.

\subsection{Annotation Challenges} 

We will now give some examples of the challenges of aligning German and Romansh words.

\subsubsection{Compound Words}
\begin{figure}
        \centering
\begin{subfigure}{.4\textwidth}
\centering        
        \tikzmarknode{WEBSEITE}{Webseite} 
        \vspace*{1cm}
        
        \tikzmarknode{pagina}{pagina} \tikzmarknode{d}{d'} \tikzmarknode{internet}{internet} 
    
    \begin{tikzpicture}[remember picture, overlay]
        \draw    (WEBSEITE) -- (pagina)
                (WEBSEITE) -- (internet);
    \end{tikzpicture}            
\end{subfigure}    
\begin{subfigure}{.4\textwidth}
\centering        
        \tikzmarknode{WEBSEITE}{Brandversicherung} 
        \vspace*{1cm}
        
        \tikzmarknode{pagina}{assicuranza} \tikzmarknode{d}{cunter} \tikzmarknode{internet}{fieu} 
    
    \begin{tikzpicture}[remember picture, overlay]
        \draw    (WEBSEITE) -- (pagina)
                (WEBSEITE) -- (internet);
    \end{tikzpicture}            
\end{subfigure}    
\caption{Aligning the German compounds \emph{Webseite} (\enquote{website}) and \emph{Brandversicherung} (\enquote{fire insurance}) to Romansh noun phrases. 
Only lexemes are aligned with each other. 
Romansh Prepositions are left unaligned.}
\label{fig:align-compounds}
\end{figure}
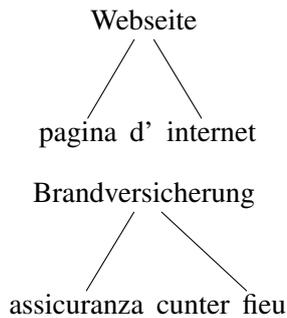

One difficulty is the alignment of German compounds (formations of new lexemes by adjoining two or more lexemes \citep{bauer1988}), to their Romansh translations, which are single words connected by a preposition (usually \emph{da}). 
We decided to align German compounds to their corresponding Romansh lexemes, leaving the Romansh prepositions unaligned. 
See Figure~\ref{fig:align-compounds} for two alignment examples.

\subsubsection{German Preterite vs.~Romansh Perfect}
In the corpus at hand, two tenses are used in German for referring to past events: the preterite and the perfect. 
The German preterite is a synthetic verb form, i.e., it is made up of a single conjugated form, e.g., \emph{nahm} (infinitive \emph{nehmen} \enquote{take}) or \emph{wurde} (infinitive \emph{werden} \enquote{become}). 
The German perfect is an analytic construction, i.e., it is made up of an auxiliary verb (\emph{haben} \enquote{have} or \emph{sein} \enquote{be}) and the past participle, e.g., \emph{Die Präsidentenkonferenz \textbf{hat} nun \textbf{entschieden}} (\enquote{the presidential conference has decided}). 

Contrary to German, Romansh only has one tense referring to past events: the perfect. 
It is an analytic construction made of, similarly to German, an auxiliary \emph{habere} \enquote{have} for transitive verbs or \emph{esse} \enquote{be} for intransitive verbs and the past participle \citep[p.~189]{bossong2008}. 

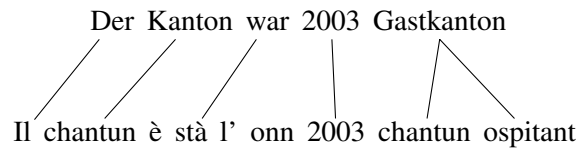
\begin{figure}
\centering        
\tikzmarknode{s0}{Der} \tikzmarknode{s1}{Kanton}  \tikzmarknode{s3}{war}  \tikzmarknode{s5}{2003} \tikzmarknode{s6}{Gastkanton}

\vspace*{1cm}

\tikzmarknode{t0}{Il} \tikzmarknode{t1}{chantun}  \tikzmarknode{t3}{è} \tikzmarknode{t4}{stà}  \tikzmarknode{t8}{l'} \tikzmarknode{t9}{onn} \tikzmarknode{t10}{2003} \tikzmarknode{t11}{chantun} \tikzmarknode{t12}{ospitant} 
\begin{tikzpicture}[remember picture, overlay, scale=0.6, every node/.style={scale=0.6}]
\draw        
(s0) -- (t0)
(s1) -- (t1)
(s3) -- (t4)
(s5) -- (t10)
(s6.south) -- (t11)
(s6.south) -- (t12)
;
\end{tikzpicture}
\caption{Alignment of German preterite to Romansh perfect. 
The German word \emph{war} is translated to Romansh \emph{è stà}. Nonetheless, \emph{è} is left unaligned since it only carries grammatical information (tense, number), but no lexical information.}
\label{fig:pret-perf}
\end{figure}

The German preterite (one word) is always translated using the Romansh perfect (two words). 
For example, in the sentence \emph{Der Kanton Graubünden war letzmals 2003 Gastkanton} (\enquote{The last time the Canton of Grisons was a host canton was in 2003}) the verb \emph{war} \enquote{was} is translated as \emph{è stà}. 
This principally results in a 1-to-2 link. 
However, since in this context Romansh \emph{è} only carries grammatical information of tense and number, but no additional lexical information, it remains unaligned, cf.~Figure~\ref{fig:pret-perf}. 

\subsubsection{Separable Verbs}
\begin{figure}
\centering        
\tikzmarknode{s0}{Regierung} \tikzmarknode{s1}{weist} \tikzmarknode{s2}{Sache} \tikzmarknode{s3}{zurück}

\vspace*{1cm}

\tikzmarknode{t0}{La} \tikzmarknode{t1}{regenza} \tikzmarknode{t2}{renvia} \tikzmarknode{t3}{la} \tikzmarknode{t4}{fatschenta}
\begin{tikzpicture}[remember picture, overlay, scale=0.6, every node/.style={scale=0.6}]
\draw        
(s0) -- (t1)
(s1) -- (t2)
(s2) -- (t4)
(s3) -- (t2)
;
\end{tikzpicture}
\caption{The German verb \emph{zurückweisen} (\enquote{reject, decline}), here separated into two words since it is used as the finite verb in the main clause, corresponds to the Romansh verb \emph{renviar}. 
This results in a 2-to-1 alignment.}
\label{fig:sep-verb}
\end{figure}
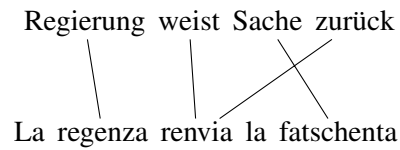

Separable verbs (German \emph{\enquote{trennbare Verben}}) are verbs in front of which affixes (mostly prepositions) are placed. 
These affixes delimit and modify the verb's meaning. In main clauses, if the finite verb is a separable verb, its affix will be placed at the end of the sentence, thus \enquote{seperable}.
Since both the verb and the affix form together the lexical meaning of the word and are conceptually inseparable, both of them should be aligned to the corresponding Romansh verb, resulting in a 2-to-1 alignment. 
See Figure~\ref{fig:sep-verb} for an example.

\subsection{Shortcomings}
Our gold standard has two shortcomings: 
First, it was created by a single annotator, the author of this paper. 
Second, it only contains Sure alignments, which might influence the evaluation (cf.~Section~\ref{subsec:evaluation}). The latter is however precedented, and there exist other word alignment gold standards which contain only Sure alignments, cf.~\citet{clematide2018} and \citet{mihalcea-pedersen-2003-evaluation}.

Refer to \citet[pp.~43-50]{dolev-mathesis-2022} for further details about the creation of the gold standard as well as more annotation examples. 

\section{Experiments and Results}\label{sec:experiments}
To test our hypothesis of whether MLMs contain information that can be meaningfully applied to Romansh, we compared two statistical word alignment systems as our baseline systems with two similarity- and embedding-based word alignment systems. 
For our baseline systems, we chose fast\_align \citep{dyer-etal-2013-simple} and eflomal \citep{Ostling2016efmaral}. 
For our similarity-based systems, we chose SimAlign \citep{jalili-sabet-etal-2020-simalign} and awesome-align \citep{dou-neubig-2021-word}.

Additionally, we tested the influence of the dataset size on the performance of the statistical models.
We also experimented with fine-tuning mBERT using awesome-align on our parallel corpus.

\subsection{Evaluation Metrics}\label{subsec:evaluation}
To evaluate the models, we consider three measurements: precision, recall and alignment error rate (AER), which are defined as follows:

\[
        \text{Recall} = \frac{|A\cap S|}{|S|},~~~~\text{Precision}  = \frac{|A\cap P|}{|A|}
\]

\[
        \text{AER} = 1- \frac{|A\cap S|+|A\cap P|}{|A|+|S|}
\]

With $A$ being the set of alignments generated by the model, $S$ being the set of Sure alignments and $P$ the set of Possible alignments, whereas $S \subseteq P$, meaning the set of Possible alignments also contains the Sure alignments \citep{och-ney-2000-improved}. In our case, $P$ equals $S$, since our gold standard only has Sure alignments, hence $P$  also stands for $S$.

To compute the measurements, we use the implementation from \citet{jalili-sabet-etal-2020-simalign}\footnotemark{}.

\footnotetext{\url{https://github.com/cisnlp/simalign/blob/master/scripts/calc_align_score.py}}

\subsection{Statistical Models}
\begin{figure}
        \centering
        \includegraphics[width=0.45\textwidth]{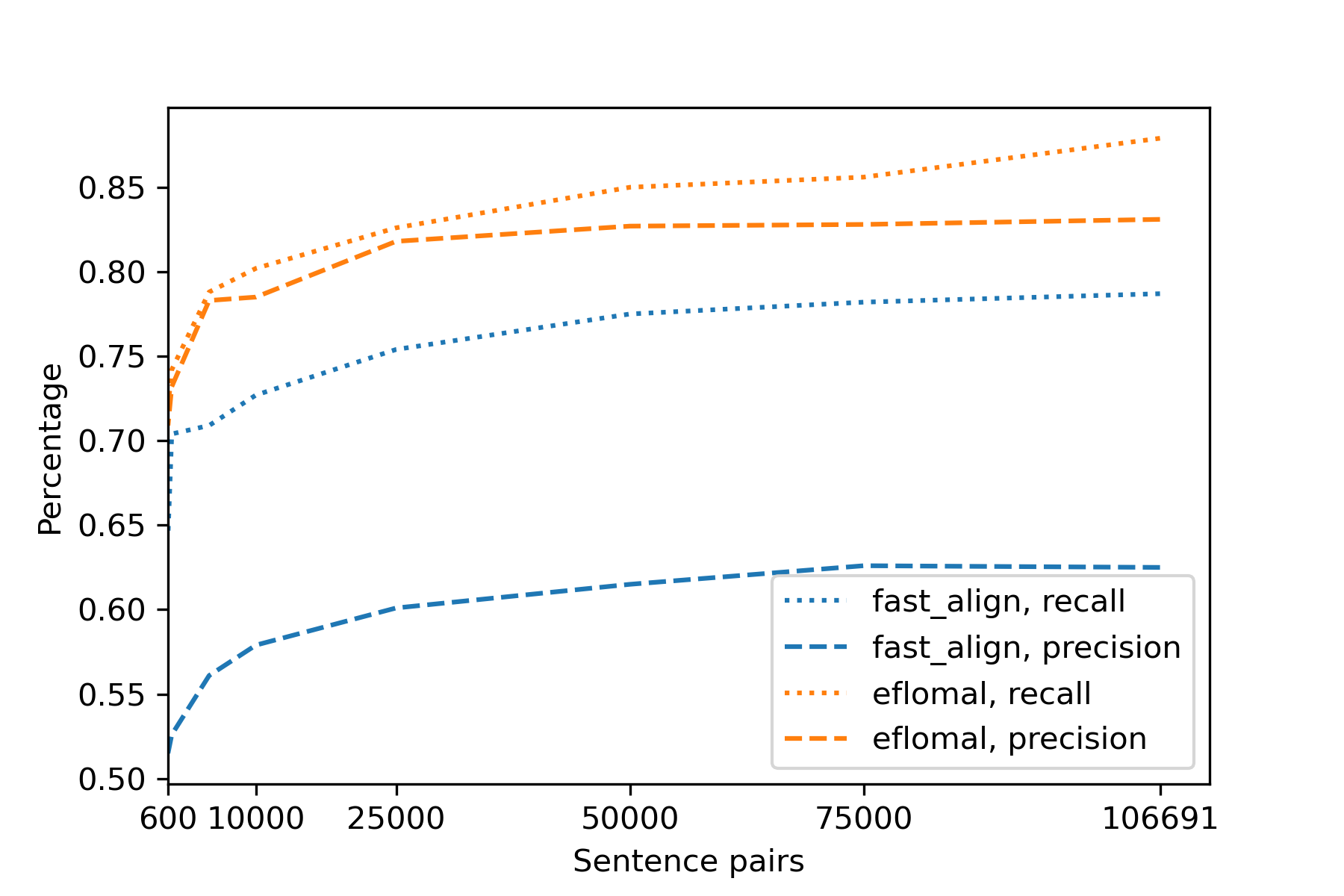}
        \caption{Performance of our baseline statistical models with relation to the dataset size.}
        \label{fig:baseline}
\end{figure}

We tested the influence of dataset sizes on the statistical models, beginning with the 600 sentences from the gold standard, and up to the entire dataset (\numprint{106091} sentences). For that, we concatenated the 600 sentence pairs from the gold standard to parts of the dataset in different sizes, trained the models, and then took the alignments for the first 600 sentences, which correspond to the annotated sentence pairs. We then compared them to the gold alignments.

\begin{figure}
\centering
\includegraphics[width=0.45\textwidth]{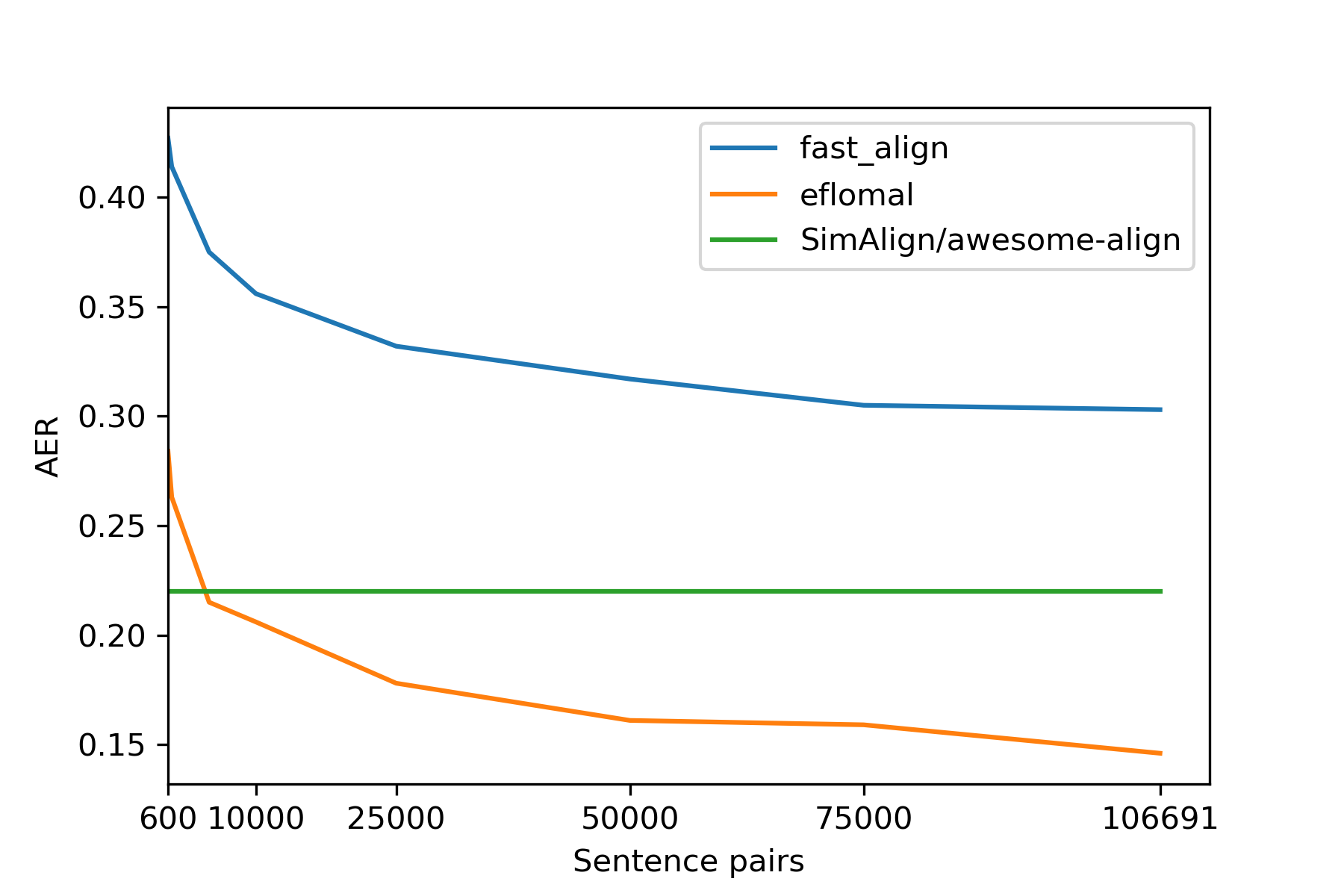}
\caption{Comparison of AER between the three systems (lower is better). The performance of fast\_align and eflomal profits from more data. The performance of SimAlign and awesome-align is not dependent on dataset size.}
\label{fig:aer}
\end{figure}

Surprisingly, fast\_align fails to match the performance that is reached by eflomal after just 600 sentences. As expected, the performance of both models increases with larger dataset sizes. Figure~\ref{fig:baseline} visualizes the changes in precision and recall with dataset sizes. Figure~\ref{fig:aer} visualizes the AER of the baseline systems compared with the similarity-based systems. The exact precision, recall and AER values are given in Table~\ref{tab:baseline} in Appendix~\ref{sec:appendix}. 

Putting these results into perspective, \citet{Ostling2016efmaral} report an AER of 0.106 for English-Swedish (\numprint{692662} sentence pairs) and an AER of 0.279 for English-Romanian (\numprint{48641}) sentence pairs (cf. Table~2 in \citet{Ostling2016efmaral}). The performance of eflomal on our German-Romansh dataset, which reaches an AER of 0.146, is within this range.

\subsection{Similarity-Based Models}
To test whether MLMs contain information that can be meaningfully applied to Romansh, we compared word alignments computed by two similarity-based alignment models, which leverage contextualized word embeddings from MLMs.
Since the dataset size does not influence the performance of similarity-based word alignment systems, we only had the 600 sentence pairs from the gold standard aligned. 

We tested the different algorithms (dubbed Argmax, Itermax and Match) offered by SimAlign to align words on a word and sub-word level and using both MLMs that are supported out-of-the-box by SimAlign: mBERT \citep{devlin-etal-2019-bert} and XLM-R \citep{conneau-etal-2020-xlm}. The embeddings were taken from the 8th layer, SimAlign's default setting. 

The results show that mBERT performs better than XLM-R, and that the Match algorithm performs best (Figure~\ref{fig:simalign-aer}; The exact precision, recall and AER values are given in Table~\ref{tab:simalign} in Appendix~\ref{sec:appendix}).

As can be seen in Figure~\ref{fig:aer} and Table~\ref{tab:fourway-comparison}, both SimAlign and awesome-align (before fine-tuning) outperform fast\_align, but are still outperformed by eflomal.

\begin{figure}
        \includegraphics[width=0.5\textwidth]{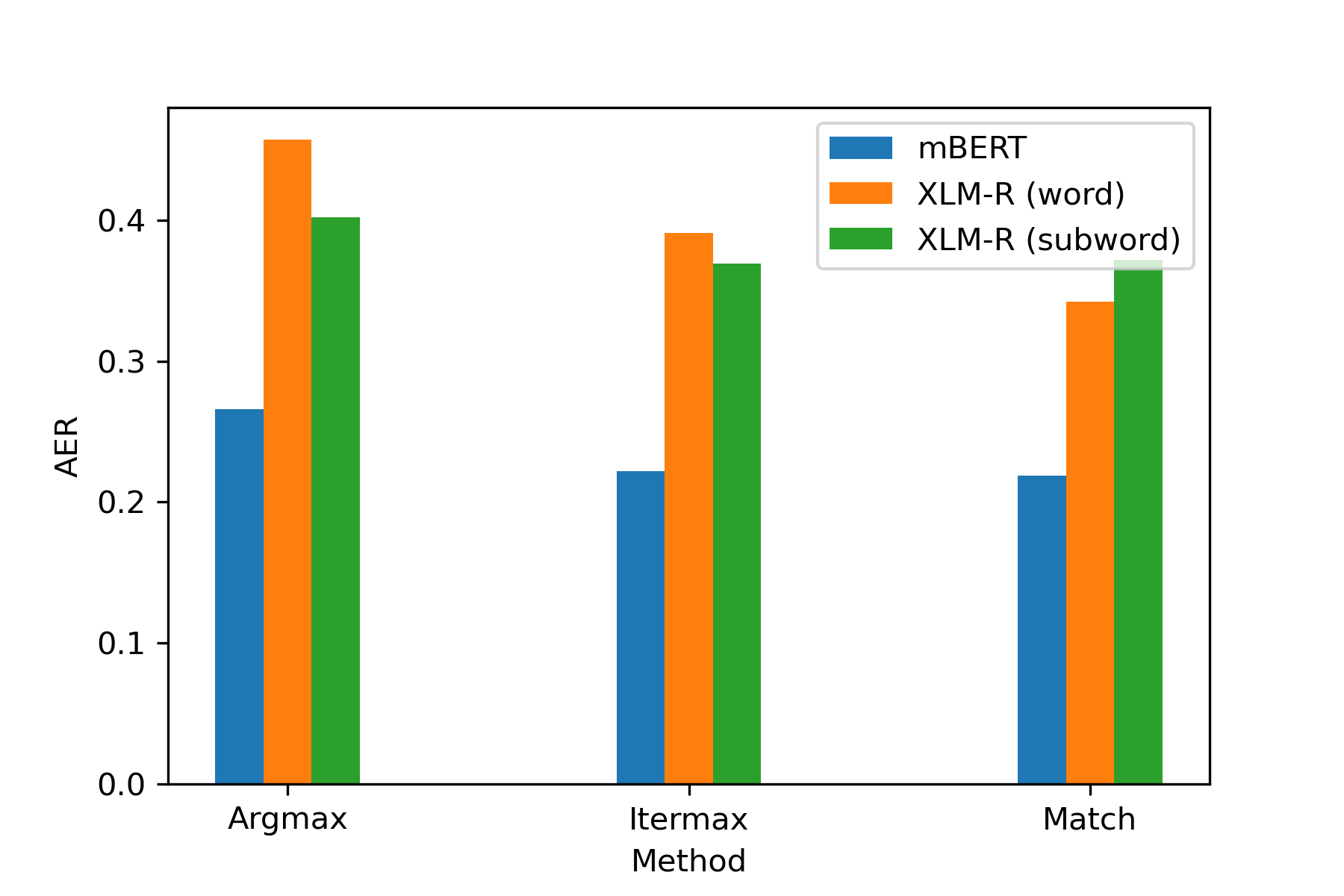}
        \caption{SimAlign's AER using different language models and settings. Argmax, Itermax and Match refer to SimAlign's algorithms for computing word alignment.}
        \label{fig:simalign-aer}
\end{figure}

To again put things into perspective, SimAlign's performance on German-Romansh sentence pairs is on par with its performance on other language pairs: For the language pairs English-German, \citet{jalili-sabet-etal-2020-simalign} report an AER of 0.19 using mBERT, which is only three percentage points better than its performance on German-Romansh. Also using mBERT, for the language pairs English-Hindi and English-Romanian, SimAlign achieves an AER of 0.45 and 0.35, respectively, which is considerably worse than its performance on German-Romansh. To conclude, SimAlign's performance is within the range of what is reported by \citet{jalili-sabet-etal-2020-simalign} for other language pairs. awesome-align performs precisely as well as SimAlign.

\subsubsection{Fine-tuning with parallel data}
awesome-align offers the option of fine-tuning an MLM on a parallel corpus for improving word alignment. 
\citet{dou-neubig-2021-word}'s fine-tuning's main objective consists of several sub-objectives. 
One sub-objective is simply continuing pre-training the language model in a masked language modeling objective on both sides of the parallel corpus. 
Another sub-objective, dubbed translation language modeling (TLM), consists of concatenating the parallel sentences and performing masked language modeling on the concatenated data. 
This objective \enquote{enable \emph{(sic)} the model to align the source and the target representations.} \cite{dou-neubig-2021-word}
Yet another sub-objective of fine-tuning, dubbed self-training objective (SO), is similar to the EM algorithm used in the IBM models. This objective brings the contextualized representations of words aligned in the first pass closer together, it reduces wrong alignments and encourages source-to-target and target-to-source alignments to be more symmetrical.

We fine-tuned mBERT using our German-Romansh parallel sentences to test the influence of fine-tuning using the settings recommended by \citet{dou-neubig-2021-word} on their GitHub page\footnote{\url{https://github.com/neulab/awesome-align}} which reportedly balance efficiency and effectiveness by applying TLM and SO.

With the fine-tuned model, awesome-align outperformed both baseline models and SimAlign, and reached an AER of 0.09. This performance lies between the AERs reported by \citet{dou-neubig-2021-word} for  German-English  (0.161) and French-English (0.041) after bilingual fine-tuning. 

Based on this big performance improvement we hypothesize that by using parallel bilingual data for fine-tuning an MLM, a previously unseen language may be easily learned by the model. Thus, MLMs could be extended to include low-resource unseen languages, by focusing on parallel data with a seen language. 

\begin{table}
        \centering
        \begin{tabular}{p{0.15\textwidth}ccc}
                \toprule
                Method & Precision & Recall & AER \\
                \midrule
                fast\_align & 0.625    & 0.787   & 0.307\\
                eflomal & 0.831 & 0.879& 0.146 \\
                \midrule
                SimAlign & 0.794  &  0.766   &   0.220 \\
                awesome-align  & 0.802   & 0.760 &  0.220 \\
                awesome-align (fine-tuned) & \textbf{0.912} & \textbf{0.911} & \textbf{0.090}\\
                \bottomrule
        \end{tabular}
        \caption{Comparison of the best results of the baseline statistical models (above) with the best results of the two embedding-based models (below). Best result in bold.}
        \label{tab:fourway-comparison}
\end{table}

\subsection{Qualtitative Evaluation}
For a short qualitative evaluation of the different models, please refer to Appendix~\ref{app:qual_eval}.

\subsection{Conclusion}
Both our conditions from Section~\ref{sec:introduction} (similarity-based alignments should be on par with statistical models, and they should be on par with performance on seen languages) were fulfilled. SimAlign's and awesome-align's performance for German-Romansh sentence pairs (1)~lies between that of our baseline models and (2)~lies within the performance range on seen languages.

We see this as evidence supporting our hypothesis that MLMs, especially mBERT, contain information that can be meaningfully applied for processing Romansh.

We also showed that mBERT reacts well to fine-tuning on German-Romansh parallel data: after fine-tuning, the AER dropped from 0.22 to 0.09.

We thus believe that leveraging MLMs, especially mBERT, would be a promising way for developing language technology and NLP applications for Romansh.

\section{Discussion}
Multilingual models such as mBERT show good performance in a so-called \enquote{cross-lingual zero-shot transfer}: fine-tuning a pre-trained model on a task, e.g., PoS tagging, on one language and then carrying out this task on a different language (target language) for which it wasn't trained \citep{deshpande-etal-2022-bert}.
Such models also perform well in a variety of tasks such as PoS tagging or NER on unseen languages, i.e., languages which were not covered by the pre-trained model, such as Faroese, Maltese or Swiss German \citep{muller-etal-2021-unseen}.

There is a lack of consensus as to what properties of a language favor performance in such scenarios, i.e., it is not entirely clear  \emph{when} zero-shot transfer works. 
Some suggest sub-word overlap is crucial for good performance \citep{wu-dredze-2019-beto}, while others show that transfer also works well between languages written in different scripts when they are typologically similar\footnotemark{}, meaning sub-word overlap is not a necessary condition \citep{pires-etal-2019-multilingual}.  It was, however, shown by  \citet{muller-etal-2021-unseen} that transliterating languages from unseen scripts leads to large gains in performance.

\footnotetext{mBERT fine-tuned for PoS tagging in Urdu (Arabic script) achieved 91\% accuracy on Hindi (Devanagari script) \citep{pires-etal-2019-multilingual}. 
Both languages are mutually intelligible and are considered variants of a single language---Hindustani.}

\citet{deshpande-etal-2022-bert} showed that zero-shot transfer is possible for different scripts with similar word order and that the lack of both, on the other hand, hurts performance. 
They also showed that zero-shot performance is correlated with alignment between word embeddings, i.e., to what extent the embeddings of different languages share the same geometric shape and are aligned across the same axes. 
\citet{muller-etal-2021-unseen} ascribe differences in mBERT's performance on unseen languages to two things: close relatedness to languages used during pretraining; and the unseen languages using the same script as those closely related languages that were seen during pretraining.

Romansh shares a high similarity, not only in script but also typologically, with other Romance languages, as well as other European languages\footnotemark{}, which are a major part of the training data for mBERT. Therefore, we interpret the good performance of SimAlign and awesome-align as a reassurance of the assumptions cited above, namely that typological similarity and sub-word overlap favor zero-shot performance.

\footnotetext{European languages from different language families (Germanic, Romance, Slavic) were shown to display high similarity to each other and to form a so-called \emph{Sprachbund}, dubbed Standard Average European \citep{haspelmath-2001-standard}.}

\section{Concluding Words}
We presented a new trilingual corpus (German, Romansh, Italian), which is a continuation of the ALLEGRA corpus by \citet{scherrer-cartoni-2012-trilingual}. We also presented a gold standard for German-Romansh word alignments, consisting of 600 sentences.

We showed that word alignments computed using similarity-based word alignment models with embeddings taken from MLMs, specifically mBERT, are on par with statistical models; they are also on par with similarity-based models' performance on seen languages. We view this as a good indication that these models contain information that is meaningful and applicable to Romansh.

\section{Ethical Considerations}
The corpus we collected consists of press releases that are publicly available on the Internet. 
The copyright lies with the Canton of Graubünden.
Downloading the press releases is explicitly allowed.

The State Chancellery of Grisons gave their approval that the data collected for this work and published therein be further available for research purposes. 
The commercial use of these documents is however prohibited.
In any event, please refer to the copyright notice\footnote{\url{https://www.gr.ch/de/Seiten/Impressum.aspx}} on the canton's website.

\subsection{Enviromental Impact}
Four hours were needed for fine-tuning mBERT for awesome-align on our GPU (nVIDIA GTX 1080). Total emissions are estimated at 0.31 kgCO$_2$eq  \footnotemark{}. Estimations were conducted using the \href{https://mlco2.github.io/impact#compute}{MachineLearning Impact calculator} presented in \citet{lacoste2019quantifying}. 100\% of the electricity used for running our machine came from renewable energy. 

\footnotetext{The equivalent of 2.5 km driven by a European car, according to the \href{https://www.europarl.europa.eu/news/de/headlines/society/20190313STO31218/co2-emissionen-von-pkw-zahlen-und-fakten-infografik}{European Parliament's website}.}

\section*{Acknowledgements}

Eyal Liron Dolev is funded by the Swiss National Science Foundation (Sinergia project grant no. 205913). I would like to thank Prof.~Martin Volk for supervising my master's thesis (cf. \citet{dolev-mathesis-2022}) which served as the basis for this paper. I would also like to thank Jannis Vamvas for encouraging me to publish this work, as well as Tannon Kew for his most valuable feedback.

\bibliography{anthology,custom}

\appendix

\section{Tables}
\label{sec:appendix}

Table~\ref{sec:corpus} gives information about the number of documents in our corpus for each language and year, as well as the number of documents that exist in at least two languages. Table~\ref{tab:baseline} provides precision, recall and AER measurements for the statistical baseline models, tested on different dataset sizes (see also Figures~\ref{fig:baseline} and \ref{fig:aer}).
Table~\ref{tab:simalign} gives information about precision, recall and AER for SimAlign's different algorithms, using different embeddings and tokenization (see also Figure~\ref{fig:simalign-aer}).

\begin{table}
        \centering
        \begin{tabular}{c|ccc|c}
                \toprule
                Year & DE & RM & IT & Parallel \\
                \midrule
                1997 & 181 & 17 & 18 & 11 \\ 
                1998 & 168 & 153 & 153 & 147 \\ 
                1999 & 161 & 130 & 130 & 131 \\ 
                2000 & 192 & 167 & 169 & 168 \\ 
                2001 & 233 & 159 & 171 & 172 \\ 
                2002 & 235 & 157 & 165 & 153 \\ 
                2003 & 167 & 110 & 111 & 102 \\ 
                2004 & 132 & 97 & 94 & 89 \\ 
                2005 & 157 & 134 & 133 & 127 \\ 
                2006 & 211 & 173 & 174 & 166 \\ 
                2007 & 199 & 147 & 145 & 140 \\ 
                2008 & 201 & 168 & 169 & 164 \\ 
                2009 & 212 & 175 & 176 & 109 \\ 
                2010 & 219 & 183 & 184 & 184 \\ 
                2011 & 203 & 167 & 167 & 167 \\ 
                2012 & 254 & 207 & 207 & 207 \\ 
                2013 & 260 & 219 & 219 & 219 \\ 
                2014 & 260 & 218 & 218 & 218 \\ 
                2015 & 227 & 183 & 183 & 183 \\ 
                2016 & 221 & 190 & 190 & 190 \\ 
                2017 & 236 & 207 & 207 & 207 \\ 
                2018 & 248 & 221 & 220 & 221 \\ 
                2019 & 238 & 216 & 216 & 216 \\ 
                2020 & 310 & 284 & 285 & 286 \\ 
                2021 & 322 & 294 & 294 & 294 \\ 
                2022 & 301 & 275 & 276 & 276 \\ 
                \midrule
                Total&\numprint{5748} & \numprint{4651}&\numprint{4674} &\numprint{4547}\\
                \bottomrule
        \end{tabular}
        \caption{Number of documents in each language per year. \enquote{Parallel} is the number of documents that exist in at least two languages.}
        \label{tab:corpus}
\end{table}

\begin{table*}[h]
        \centering
        \begin{tabular}{ccccc}
        \toprule
                              Method &Dataset Size & Precision & Recall    & AER \\
        \midrule 
        \multirow{8}{*}{fast\_align} 
        & 600                   & 0.515   & 0.644  & 0.427 \\
                            
                             & 1k                    & 0.526   & 0.662  & 0.414 \\
                             & 5k                    & 0.561   & 0.704  & 0.375 \\
                             & 10k                   & 0.579   & 0.725  & 0.356 \\
                             & 25k                   & 0.601   & 0.75   & 0.332 \\
                             & 50k                   & 0.615   & 0.768  & 0.317  \\ 
                            & 75k                  & 0.626    & 0.781  & 0.305 \\  
                              &  \numprint{100}k   & \textbf{0.625}    & \textbf{0.787}   & \textbf{0.307} \\
                             \cmidrule{1-5}
                              \multirow{8}{*}{eflomal} 
                              & 600              & 0.707 & 0.724  & 0.284\\
                              &           1k    & 0.732 & 0.742 & 0.263 \\
                              &           5k    & 0.776 & 0.78 & 0.222\\   
                              &           10k   & 0.798 & 0.805  & 0.199 \\
                              &           25k   & 0.812  &0.836  & 0.176 \\

                              &           50k    & 0.828 & 0.85  & 0.161 \\
                              &           75k & 0.827 & 0.856 & 0.159 \\
                              & \numprint{100}k & \textbf{0.831} & \textbf{0.879} & \textbf{0.146} \\
        \bottomrule
        \end{tabular}
        \caption{Word alignment quality of the baseline statistical models, tested on different dataset sizes. 
  Best result per method in bold.
  \enquote{Dataset Size} refers to the number of sentence pairs. 
  The full dataset size  is the number of sentence pairs extracted at the time of the experiments 
  (\numprint{106691}) plus the 600 annotated sentence pairs from the gold standard.}
        \label{tab:baseline}
\end{table*}

\begin{table*}[h]
        \centering
        \begin{tabular}{llllccc}
        \toprule
         &	 Model	     & Level		              & Method & Precision & Recall    & AER \\
        \midrule
        \multirow{9}{1em}{\rotatebox{90}{SimAlign}} & \multirow{3}{*}{mBERT} & \multirow{3}{*}{Subword}  &  Argmax & \textbf{0.894}    & 0.622  & 0.266 \\
         &	&     &  Itermax & 0.832  & 0.731  & 0.222 \\
            & &	 &  Match   & 0.795  & \textbf{0.781}  & \textbf{0.220} \\	
        \cmidrule{2-7}
        & \multirow{6}{*}{XLM-R} & \multirow{3}{*}{Word} &  Argmax  & \textbf{0.848} & 0.399  & 0.457 \\
        &	&  & Itermax  & 0.767  & 0.504  &  0.391 \\
        &	&    & Match    & 0.67 & 0.647 & \textbf{0.342} \\                                                                      \cmidrule{3-7}
         &  & \multirow{3}{*}{Subword}	 & Argmax  & 0.773  & 0.488  & 0.402 \\
        &  & & Itermax                   & 0.671    & 0.595  & 0.369 \\
        &  &	& Match	        	& 0.558  & \textbf{0.719}   & 0.372 \\      
        \bottomrule
        \end{tabular}
        \caption{Word alignment quality using SimAlign, with embeddings taken from different MLMs using word/sub-word level. Embeddings were taken from the eighth layer (default SimAlign setting).
        Best result per embedding type in bold.}
        \label{tab:simalign}
\end{table*}

\section{Qualitative Evaluation}
\label{app:qual_eval}
We attempt to give a short qualitative evaluation of the advantages and shortcomings of the models we evaluated, i.e., describe which linguistic features the models deal with well and which features make them fail. To this end, we chose three sample sentences containing two potential pitfalls: aligning compound words and aligning differing word order. We compare our annotations, i.e., the gold standard, with eflomal, the best statistical model, with SimAlign (using embeddings from mBERT and the Match method) and with awesome-align, before and after fine-tuning.

In the following plots, the dark-colored squares (green/blue) represent the gold alignment. The green plots compare eflomal and SimAlign. The blue plots compare awesome-align before and after fine-tuning.
The plots were created using a slightly altered version of a script made available by \citet{jalili-sabet-etal-2020-simalign} on their GitHub page\footnote{\url{https://github.com/cisnlp/simalign/blob/master/scripts/visualize.py}}.

\begin{figure*}
        \includegraphics[width=0.49\linewidth]{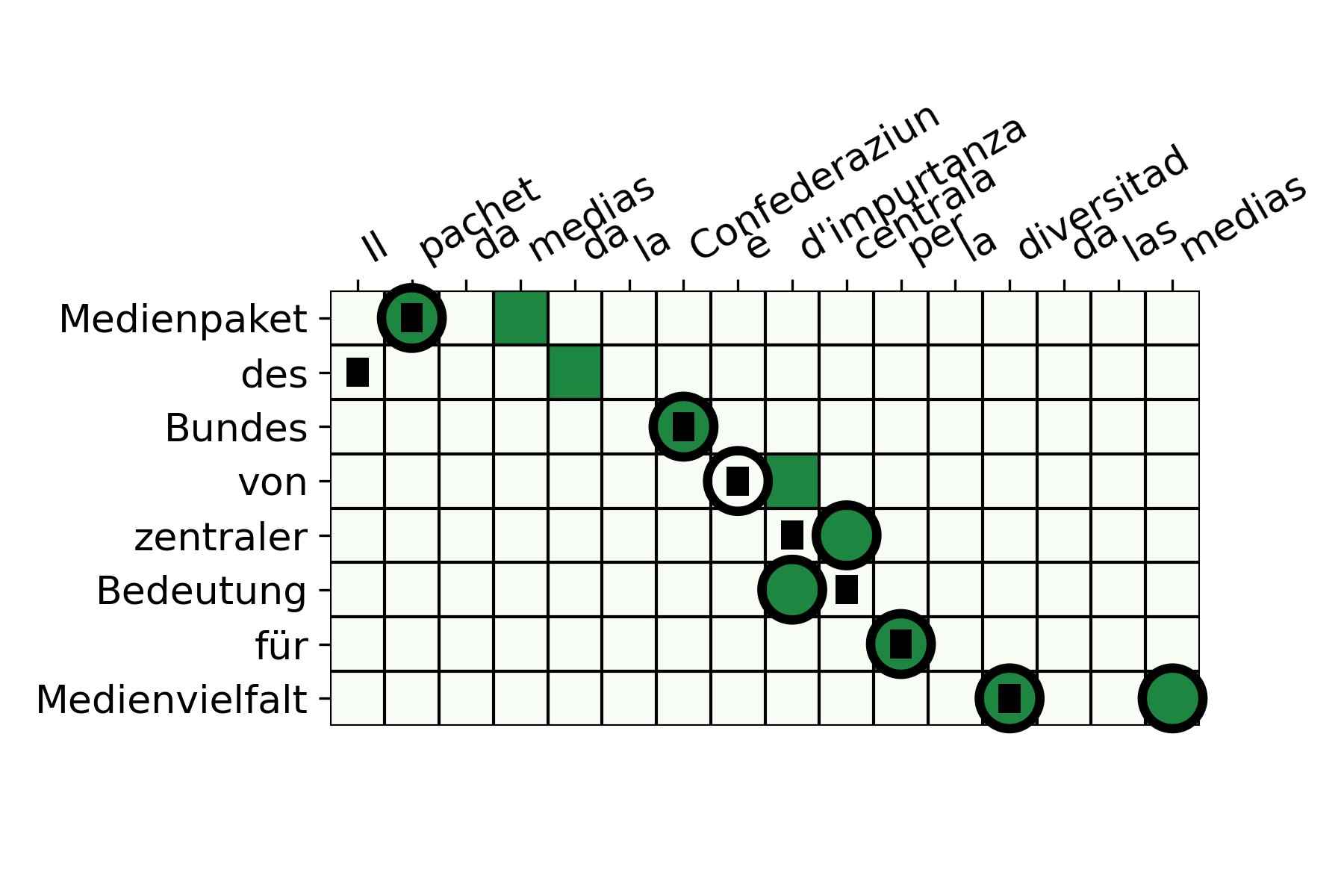}
        \includegraphics[width=0.49\linewidth]{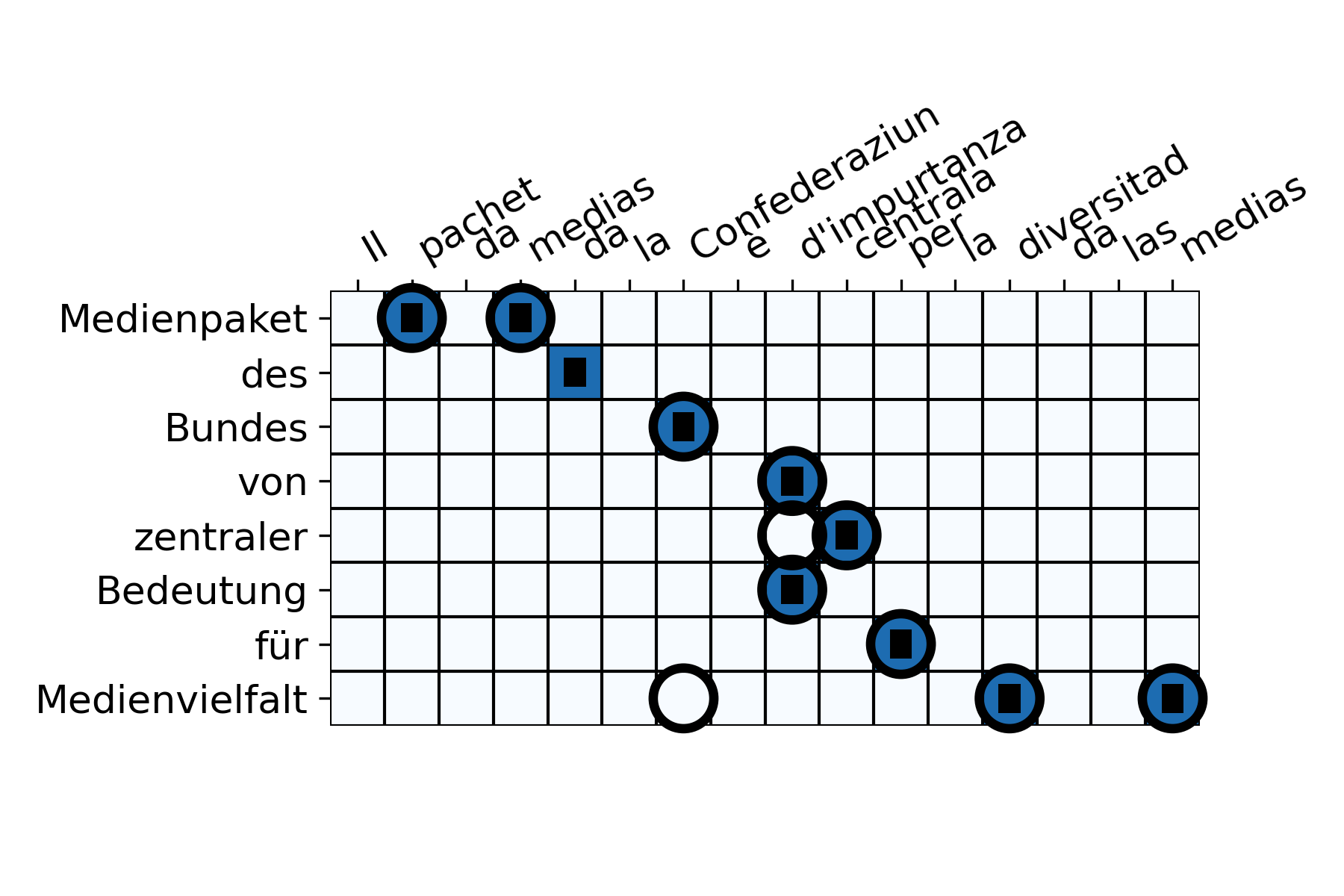}
        \caption{Example for aligning compounds. Green: eflomal (boxes) vs. SimAlign (circles); Blue: awesome-align (circles) vs. after fine-tuning (boxes). The dark green/blue squares represent the gold alignments.}
        \label{fig:s88}
\end{figure*}

It seems that of all systems, SimAlign has the biggest difficulty aligning German compounds to their Romansh counterparts. In Figure~\ref{fig:s88}, SimAlign (green, boxes), fails to align the words \emph{Medienpaket} and \emph{Medienvielfalt}, whereas eflomal (green, circles) manages to align at least \emph{Medienvielfalt} successfully. awesome-align (blue, circles) manages to align both compounds correctly but still has some precision problems. After fine-tuning (blue, boxes), awesome-align's accuracy is perfect.

\begin{figure*}
  \includegraphics[width=0.49\linewidth]{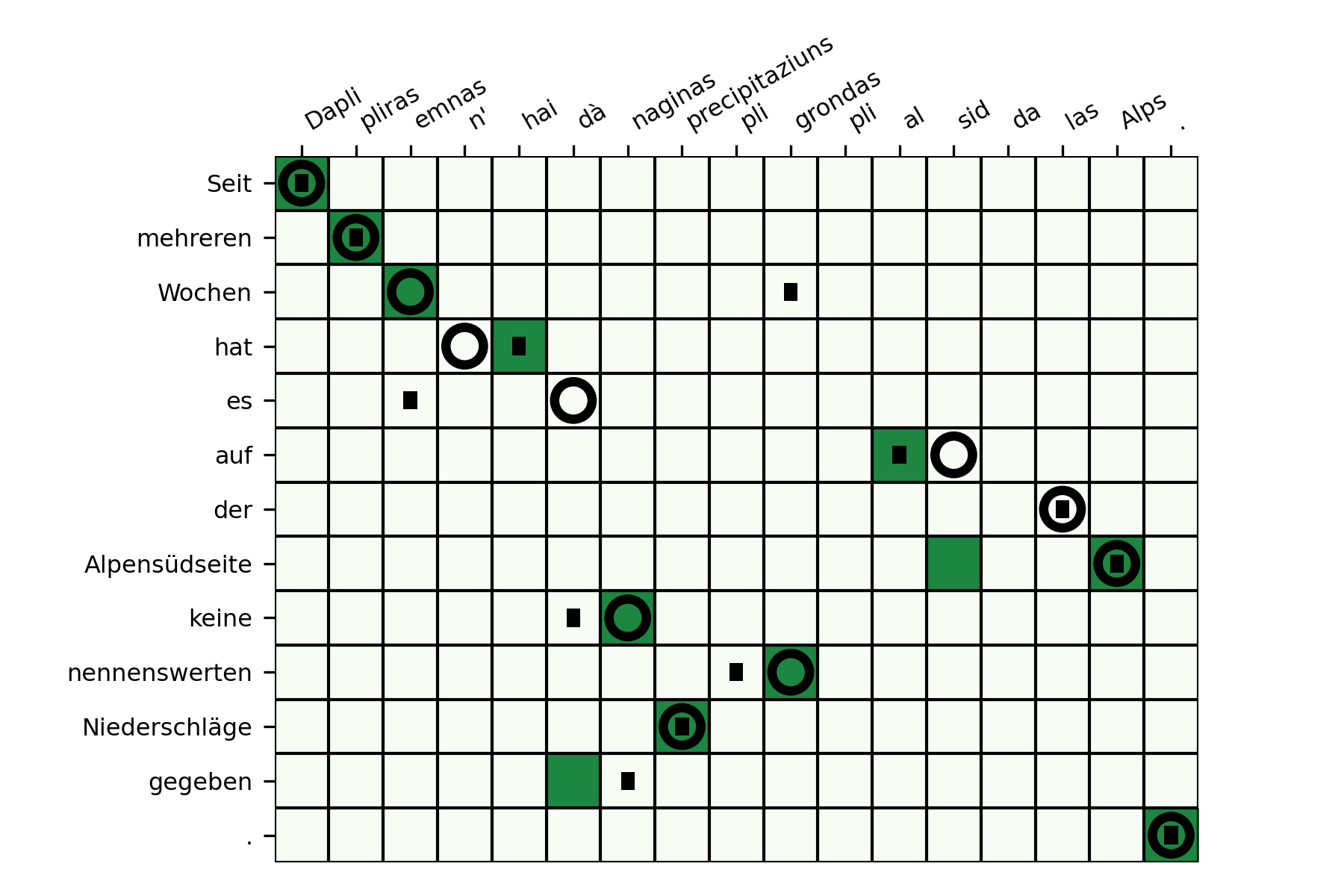}
  \includegraphics[width=0.49\linewidth]{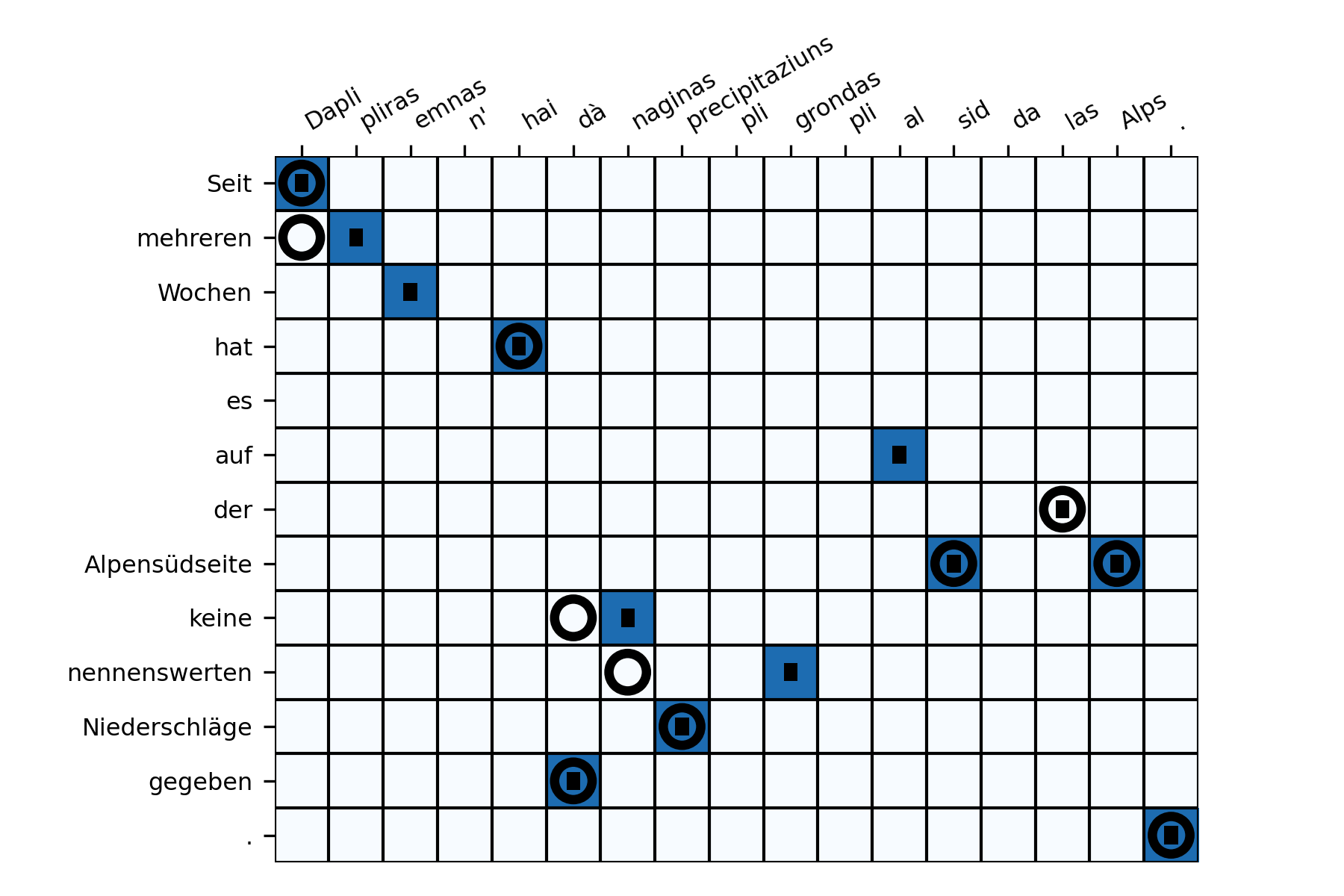}
  \caption{Example for aligning a sentence with differing word order. Green: eflomal (boxes) vs. SimAlign (circles); Blue: awesome-align (circles) vs. after fine-tuning (boxes). The dark green/blue squares represent the gold alignments.}
  \label{fig:s196}
\end{figure*}

The sentence in Figure~\ref{fig:s196} is an example of differing word order between German and Romansh. All models seem to have difficulties aligning this sentence. eflomal (green, circles) performs best with an accuracy of 0.66, followed by awesome-align (blue, circles) and SimAlign (green, boxes), with an accuracy of 0.61 and 0.58, respectively. After fine-tuning, awesome-align (blue, boxes), outperforms the other models with an accuracy of 0.92. 

\begin{figure*}
        \includegraphics[width=0.49\linewidth]{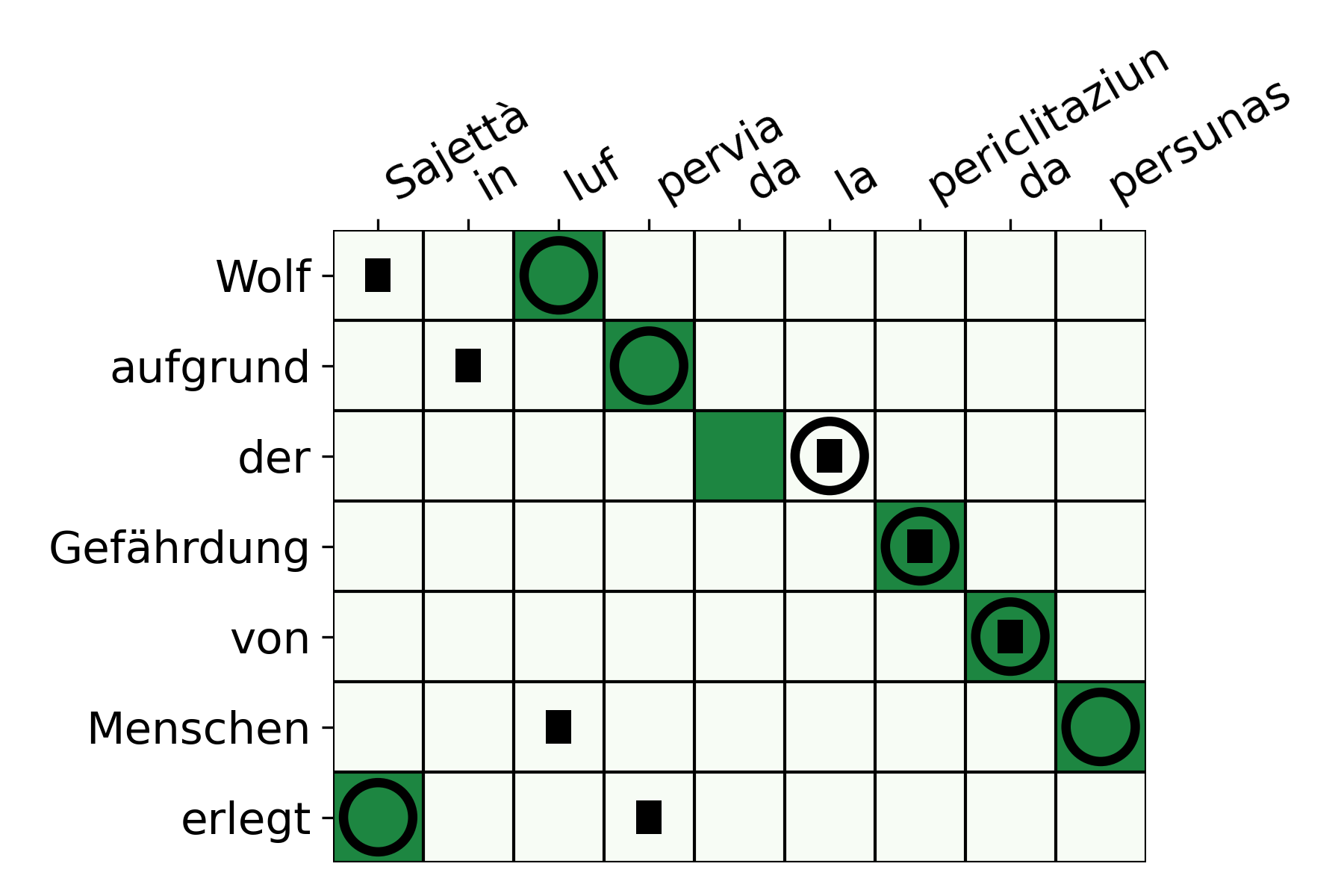}
        \includegraphics[width=0.49\linewidth]{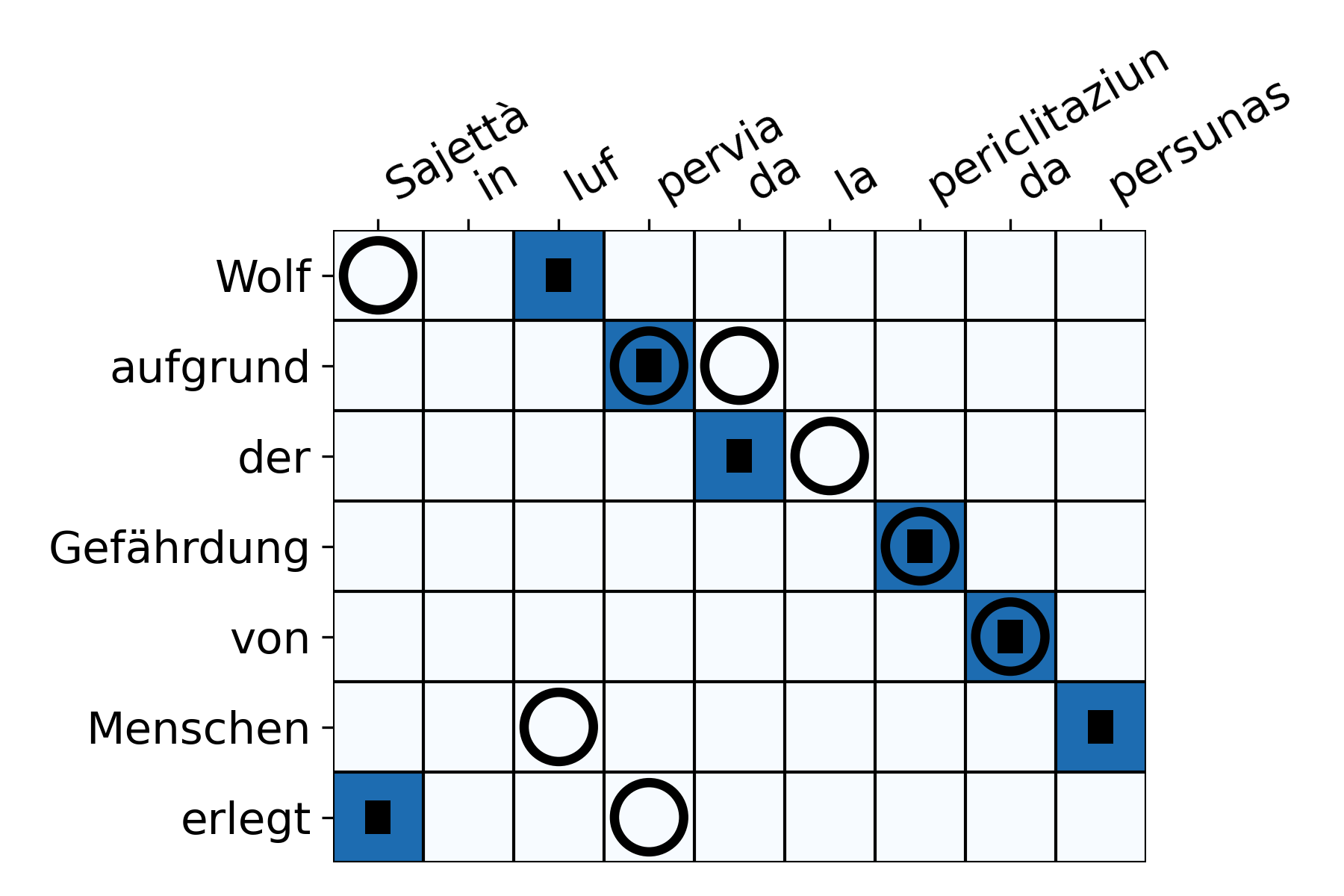}
        \caption{Example for the clear improvement after fine-tuning. Green: eflomal (boxes) vs. SimAlign (circles); Blue: awesome-align (circles) vs. after fine-tuning (boxes). The dark green/blue squares represent the gold alignments.}
        \label{fig:s90}
\end{figure*}

The sentence in Figure~\ref{fig:s90} is a general example of an unsuccessful alignment: Both SimAlign and awesome-align fail, although it is not quite clear why (presumably too many Romansh words are not similar enough to words in seen languages). SimAlign (green, boxes) and awesome-align (blue, circles) have a very low accuracy of 0.33 and 0.4, respectively. eflomal, the statistical model (green, circles), reaches an accuracy of 0.86. However, after fine-tuning, awesome-align (blue, boxes) seems to have learned the right word correspondences and reaches perfect accuracy. 

To conclude, a superficial and unsystematic comparison between the systems gives the impression that both SimAlign and awesome-align have difficulties aligning compounds and differing word order. 
Romansh words that are less similar to words in seen languages might also have a negative impact on this. 
As the case may be, fine-tuning mBERT using awesome-align's training objectives helps awesome-align deal with those difficulties and shows a big improvement in accuracy.

\end{document}